\pdfoutput=1

\documentclass[11pt]{article}

\usepackage[preprint]{acl}

\usepackage{stfloats} 
\usepackage{times}
\usepackage{latexsym}

\usepackage[T1]{fontenc}

\usepackage[utf8]{inputenc}

\usepackage{microtype}

\usepackage{inconsolata}

\usepackage{graphicx}
\usepackage{subcaption}
\usepackage{booktabs}
\usepackage{graphicx}
\usepackage{float}
\usepackage{tcolorbox}
\usepackage{listings}
\usepackage{tabularx}
\usepackage{booktabs}
\usepackage{enumitem}
\setlist{nosep}
\usepackage{cuted}

\title{Taming the Real-world Complexities in CPT E/M Coding with Large Language Models}

\author{Islam Nassar, Yang Lin, Yuan Jin, Rongxin Zhu, Chang Wei Tan, Zenan Zhai, \\\textbf{Nitika Mathur}, \textbf{Thanh Tien Vu}, \textbf{Xu Zhong}, \textbf{Long Duong}, \textbf{Yuan-Fang Li} \\
Oracle Health \& AI\\
\tt\small\{islam.nassar, yang.y.lin, yuan.j.jin, rongxin.zhu, changwei.tan, zenan.zhai, nitika.mathur, \\\tt\small thanh.v.vu, peter.zhong, long.duong, yuanfang.li\}@oracle.com
}

\begin{document}
\maketitle
\begin{abstract}
Evaluation and Management (E/M) coding, under the Current Procedural Terminology (CPT) taxonomy, documents medical services provided to patients by physicians. Used primarily for billing purposes, it is in physicians' best interest to provide accurate CPT E/M codes. 
Automating this coding task will help alleviate physicians' documentation burden, improve billing efficiency, and ultimately enable better patient care. However, a number of real-world complexities have made E/M encoding automation a challenging task. In this paper, we elaborate some of the key complexities and present ProFees, our LLM-based framework that tackles them, followed by a systematic evaluation. On an expert-curated real-world dataset, ProFees achieves an increase in coding accuracy of more than 36\% over a commercial CPT E/M coding system and almost 5\% over our strongest single-prompt baseline, demonstrating its effectiveness in addressing the real-world complexities. 
\end{abstract}

\section{Introduction}

Accurate and efficient coding of Evaluation and Management (E/M) services using the Current Procedural Terminology\footnote{\url{https://www.ama-assn.org/practice-management/cpt/cpt-overview-and-code-approval}} (CPT) system is crucial for healthcare providers, as it directly impacts billing accuracy, regulatory compliance, and revenue cycles. Traditionally, CPT E/M coding is performed by physicians and trained human coders who review clinical encounter notes and electronic health record (EHR) data to assign appropriate codes. They often rely on detailed guidelines~\cite{ama2024cpt} to navigate the complexity of Medical Decision Making (MDM) for a given encounter. However, this manual process is resource-intensive, prone to inconsistencies due to variation in coder's expertise, and vulnerable to errors arising from intricate coding guidelines and the nuanced nature of clinical reasoning.

Many studies highlight the financial and compliance ramifications of coding inaccuracies. For example, the U.S. Office of Inspector General (OIG) estimated that in 2010, Medicare inappropriately paid US\$6.7 billion (21\% of total E/M payments) due to incorrect coding and/or insufficient documentation; 42\% of claims were miscoded and 19\% lacked proper documentation~\cite{oig2014em}. A Florida-specific analysis estimated that nearly 9\% of primary care visits (2.6 million annually) were undercoded, resulting in approximately US\$114 million in lost revenue for hospitals~\cite{tenpas2023florida}. These findings underscore both widespread inconsistency and the high cost of coding errors.

Automation of CPT E/M coding offers several compelling advantages, including reducing human error through consistent application of complex coding rules, achieving scalable efficiency by handling high-volume chart reviews without increasing workload on physicians and coders,  enhancing timely revenue capture by improved turnaround time. Additionally, automated coding can provide transparent and traceable reasoning, crucial for audits and compliance risk mitigation and enhance coding consistency, and optimizes coder resources by allowing human coders to focus on more complex or ambiguous cases. Beyond its immediate healthcare impact, CPT E/M coding exemplifies a broader class of problems where models must map free-text to codified decisions under strict guideline constraints. Insights from this work therefore extend to other regulated domains that require auditable, rule-aligned predictions.


\begin{table*}[t]
\small
\centering
\caption{Key challenges in CPT E/M automation and our corresponding solutions.}
\label{tab:challenge_solution}
\begin{tabular}{@{}p{0.1\linewidth} p{0.40\linewidth} p{0.45\linewidth}@{}}
\toprule
\textbf{Challenge} & \textbf{Description} & \textbf{Solution} \\
\midrule
Intermediate labels & Production data include only the final CPT code, while the underlying medical decision-making (MDM) level with 3 intermediate element levels are absent. & 117 encounters are re-annotated by our internal expert with the 3 intermediate elements -- Problem, Data, and Risk levels + justifications. \\
\midrule
Label noise & Even expert coders disagree: in our case, external and internal coding experts diverge on 56\% of encounters (Sec.~\ref{sec:data}). & The development data is split into \textit{Platinum} (agreement) and \textit{Disagreement}; We first tune our prompts on \textit{Platinum}, then iteratively refine using error analysis on both subsets with the internal expert. \\
\midrule
Explainability &  Clinicians and auditors require human-readable reasoning and black-box predictions are unacceptable. & Chain-of-thought exemplars and checklist-based critic outputs accompany every prediction, explicitly citing CPT guideline clauses. \\
\midrule
Robustness &A production system must deliver repeatable results. LLM-based method could be potential solution but outputs could be stochastic. & Self-consistency ensemble (\(K{=}3\)) with majority voting plus a deterministic decision tree ensures repeatable results at controlled cost. \\
\midrule
Clinical breadth &Correct coding demands deep, wide-ranging familiarity with guidelines and edge cases. & Dynamic few-shot retrieval injects guideline-aligned examples; iterative error analysis and prompt tuning add edge-case expertise. \\
\bottomrule
\end{tabular}
\end{table*}

In this paper, we share our practical experience in developing and deploying ProFees, an automated CPT E/M coding model within a production EHR system. Development of ProFees provides solutions to five main complexities of real-world E/M coding, summarized in Table \ref{tab:challenge_solution}. Leveraging advanced large language model (LLM) techniques, ProFees employs a dynamic few-shot prompting strategy to align coding decisions with best practices contextually relevant to each patients' encounter notes. Furthermore, an LLM-based critic is integrated to validate reasoning across the MDM elements explicitly, ensuring robust and explainable outcomes. To mitigate the stochasticity inherent in generative models, ProFees also adopts a self-consistency strategy to achieve reliable and consistent coding predictions.


This paper's key contributions include:
\begin{itemize}
\item Development and deployment of ProFees, a sophisticated LLM-based framework employing dynamic few-shot prompting, explicit criticism, and self-consistency techniques for accurate CPT E/M coding and intermediate MDM level estimation.
\item Empirical demonstration of ProFees's efficacy on a proprietary test dataset\footnote{This test set was meticulously annotated by our internal expert in the same way as the development set of 117 encounters. Due to privacy and regulatory constraints, this test set is not publicly available.} with comparison against a commercial CPT E/M coding system and single-prompt baselines. Our findings highlight superiority of ProFees in E/M coding accuracy over both the commerical system and our strongest baseline by around 36\% and 5\% respectively, validating its efficacy in CPT E/M coding automation.
\end{itemize}

\begin{figure*}[t]
  \centering
  \includegraphics[width=\textwidth]{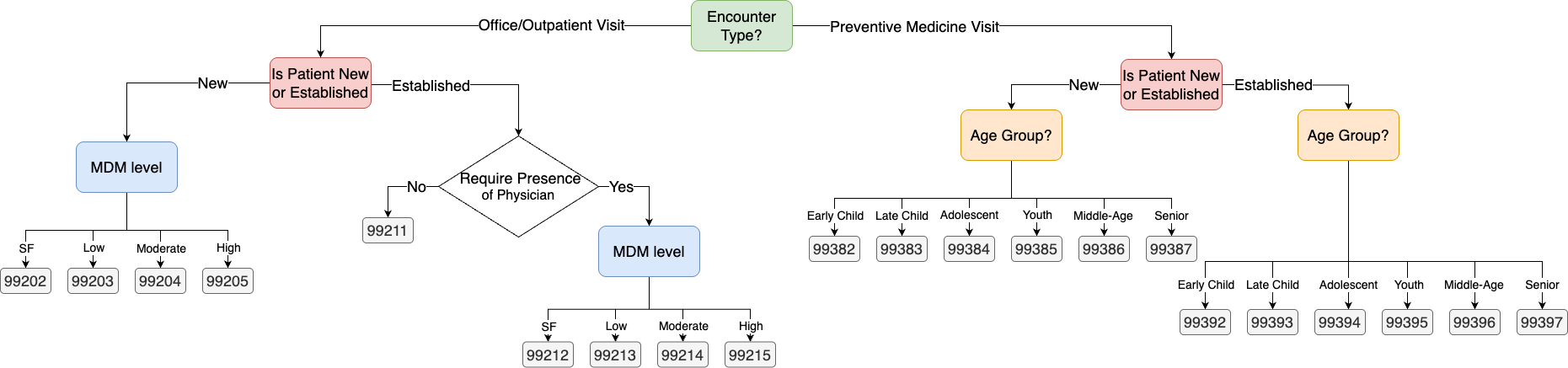}
  \caption{CPT E/M coding decision tree used to guide human coding of office and preventive visits. Other visit types are omitted for brevity. ``SF'' stands for ``straightforward''.}
  \label{fig:decision_tree}
\end{figure*}

\section{Related Work}

CPT coding has gone through three broad phases of modeling and technical development, which are classical machine learning, deep learning and large language modeling. 

Classical ML models heavily rely on feature engineering, which demands significant pre-processing efforts to convert structured and/or unstructured text data into hand-crafted features. Supervised classifiers, such as support-vector machines and gradient-boosted trees, are then learned on these features to predict CPT codes~\cite{Morey2025, Khaleghi2021}. While achieving moderate accuracy, these models struggle with the complexity and variability of clinical narratives.

Deep learning models for CPT coding mainly feature encoder-decoder architectures. The encoder learns to map medical text into dense embeddings, based on which the decoder learns to predict the codes. Despite the wide success of DL models in ICD-10 coding~\cite{Ji2024}, their application to CPT coding is limited and has not been reported with favorable results compared to classical ML models~\cite{Burns2020, Levy2022, Kim2023}, which is likely due to the unavailability of large datasets for effective training.

Recently, LLMs have broken new grounds in medical coding, where with prompt engineering, GPT-4 is reported to have achieved comparable performance to SOTA DL models for ICD-10 coding~\cite{boyle2023automated}. However, other studies also found that na\"ively feeding raw texts and code descriptions into LLMs for code generation can only yield sub-optimal results that fall short of the coding performance of human coders~\cite{soroush2024llmcoding}. Overall, the development of effective LLM-based frameworks for medical coding is still in its infancy. This paper presents the first systematic attempt to develop such a framework for CPT E/M coding.
\section{Task Overview}

\textbf{CPT E/M coding} involves assigning standardized billing codes based on the type and complexity of the patient encounter. In this work, we focus on outpatient and preventive medicine encounters. Figure~\ref{fig:decision_tree} illustrates the decision process used by human coders, starting from determining the encounter type, followed by key branching logic such as whether the patient is new or established, and finally selecting the appropriate CPT code based on either the Medical Decision Making (MDM) level or the patient age group.

Among these branches, determining the MDM level is the most complex and error-prone step. MDM is assessed across three elements: (1) the number and complexity of \emph{problems} addressed during patient visit, (2) the amount and/or complexity of \emph{data} reviewed and analyzed, and (3) the \emph{risk} of complications and/or morbidity or mortality. The final MDM level is determined using a \emph{2-out-of-3} rule, that at least two of the three elements must meet the criteria for a given complexity level (i.e. ``straightforward'', ``low'', ``moderate'', or ``high''). 

This process requires structured clinical reasoning. For instance, for the ``problems addressed'' element, distinguishing between ``stable, chronic illness'' (i.e. ``low'' complexity) and ``chronic illnesses with exacerbation, progression, or side effects of treatment'' (i.e. ``moderate'' complexity) depends on close reading of clinical descriptions and inferred disease trajectories. Similarly, assessing the ``data complexity'' element may involve identifying whether documentation reflects ``ordering unique test(s)'' or ``independent interpretation of tests'', which can elevate the data complexity from ``low'' to ``moderate'' or ``extensive''. See Appendix~\ref{sec:mdm_guidelines} for a full view of the 2024 edition of MDM guidelines used for developing ProFees 
and Appendix~\ref{sec: synthetic example} for a synthetic example.

In this paper, we report only strict exact-match accuracy, as payers adjudicate at the code level and incorrect CPT E/M codes may be denied, downcoded, or adjusted, affecting reimbursement timelines. Near-miss codes do not change clinical care (E/M is a billing construct), but they have operational and financial implications: miscoding can trigger denials and rework; systematic overcoding increases legal/compliance risk despite any short-term gain; systematic undercoding reduces revenue and understates provider workload. In addition, although business outcomes, e.g., revenue uplift, are ultimately important, they are much harder to be attributed directly to CPT E/M code choices due to other down-the-line factors, such as diagnosis–procedure linkage, billing completeness, payer-specific rules, and documentation quality. We therefore focus our evaluation on exact-match accuracy.



\section{The ProFees Model}

\begin{figure}[t]
\centering
\includegraphics[width=.45\textwidth]{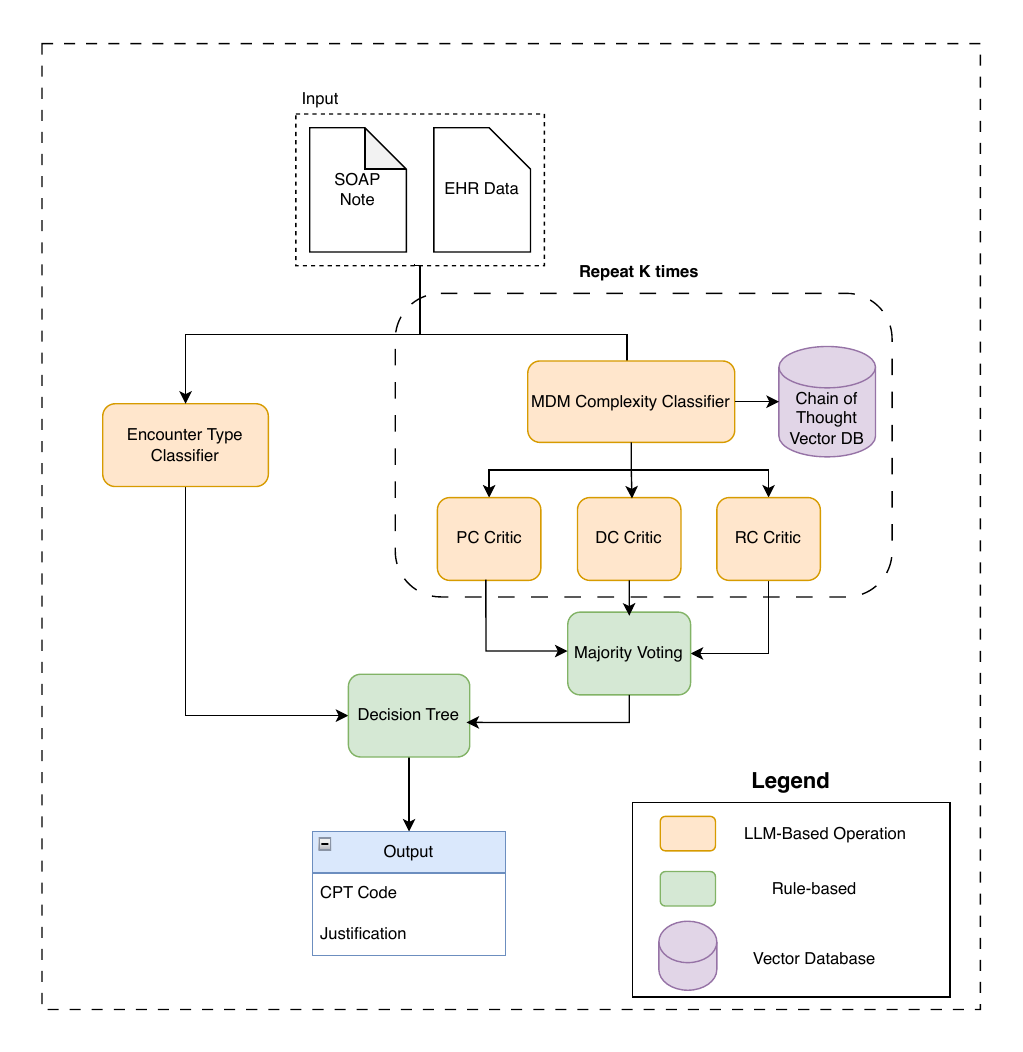}
\caption{ProFees architecture for CPT E/M coding prediction. PC, DC and RC stand for problem, data and risk complexity respectively.}
\label{fig:model_architecture}
\end{figure}


As illustrated in Fig.~\ref{fig:model_architecture}, our proposed ProFees model for CPT E/M coding integrates LLM-based classifiers and self-critics, few-shot in-context learning \cite{brown2020languagemodelsfewshotlearners} boosted by retrieval with chain-of-thought \cite{wei2023chainofthoughtpromptingelicitsreasoning} exemplars, and traditional rule-based decision trees. The architecture comprises two primary LLM-driven modules: the \textit{Encounter Type Classifier}, which identifies the nature of the clinical visit (e.g., inpatient, outpatient and preventive medicine), and the \textit{MDM Complexity Classifier}, which predicts the level of medical decision-making (MDM) involved.

This modular decomposition is directly motivated by CPT E/M coding guidelines, which determine the final code based on three key inputs: encounter type, MDM level, and patient type, with the patient type being readily available from structured EHR data. Therefore, our model is designed to predict the two remaining components: encounter type and MDM level. The initial MDM prediction is further refined through a Recursive Criticism and Improvement (RCI) process~\cite{kim2023language} to ensure guideline compliance. The outputs of both classifiers are then combined and processed by a downstream rule-based decision tree to produce the final CPT E/M code together with the corresponding justification.

\subsection{Design Rationale and Practical Impact}
Our model design reflects iterative failure analysis in collaboration with medical coders, addressing limitations of early monolithic prompt architectures that lacked interpretability, guideline adherence, and reliability. By modularizing prediction, reasoning, and validation, the system ensures contextual relevance through retrieval-augmented prompting, enforces compliance via recursive self-critique, and reduces hallucination by rule-based logic (e.g., majority voting, 2-out-of-3 CPT rules) to deterministic decision trees. This architecture achieves high accuracy, robust explainability via CoT and critic reviews, and strong alignment with clinical workflows—supporting trust, adoption, and real-world deployment.



\subsection{Encounter Type Classifier}
This component identifies the type of encounters using an LLM. The model receives relevant contextual information such as SOAP notes\footnote{SOAP notes are a highly structured and standard format for documenting patient encounters, containing the following parts: \textbf{S}ubjective, \textbf{O}bjective, \textbf{A}ssessment and \textbf{P}lan. See Appendix~\ref{sec: synthetic example} for an example.}, patient age, and patient type, and returns an appropriate encounter classification.

\subsection{MDM Complexity Classifier}
This component assesses the complexity level involved in medical decision-making. The prediction occurs in two stages: an initial classification followed by self-refinement through RCI. The initial prediction leverages dynamic few-shot prompting with external annotated examples retrieved from a Vector Database (VDB), the content of which consist of justifications and reasoning steps that were curated and validated by human coding experts to align with MDM guidelines. Subsequently, the RCI process systematically audits and refines the initial prediction to ensure adherence to comprehensive guideline criteria. To further enhance robustness, the model employs self-consistency via majority voting across multiple LLM inferences.

\subsubsection{Dynamic Few-Shot CoT Prompting}\label{sec:dynamic_few_shot} 
To enable the MDM classifier to leverage annotated external knowledge for reference in reasoning and decision making. We designed and created a VDB that indexes encounters of the development and test datasets described in Section~\ref{sec:data}, each consisting of:
\begin{itemize}
    \item \textbf{Gold justifications} per MDM element, provided by our internal coding experts.
    \item \textbf{Model justifications} per MDM element, generated by an LLM, curated and verified to be correct by our internal experts.
    \item \textbf{Chain-of-thought reasoning} (with step-by-step verification on well-designed checklists) for each MDM element, generated by an LLM, curated and verified by our internal experts.
\end{itemize}

During retrieval, the SOAP note of an input encounter is used to query the VDB to select Top-$N$ relevant exemplars using semantic vector search.

\subsubsection{Recursive Criticism and Improvement}
Despite the careful prompt engineering to align the MDM classifier with the CPT coding guidelines, we found that it still remained limited in cases that involve criteria not explicitly addressed by the guidelines or require clinical knowledge beyond the current capabilities of LLMs.

To overcome these limitations, we introduced additional self-critics to follow the initial MDM classification for all the MDM elements, i.e., the ``problem'' and ``data'' complexity, and ``risk of complications''. Each of the critics is prompted to review and critically evaluate the intermediate predictions of the corresponding MDM elements, by following checklists designed to address common generic errors identified by our internal expert for each MDM element.

\subsubsection{Self-Consistency via Majority Voting}\label{sec:self_consistency} 
A key challenge with LLMs is their inherent stochasticity; that even with identical inputs and controlled conditions, predictions can vary. As discussed by \citeauthor{Atil2025NonDeterminism}~(\citeyear{Atil2025NonDeterminism}), even with fixed random seeds and identical model settings, deterministic model outputs cannot be guaranteed due to factors such as input subjectivity, variability in the LLM backend infrastructure, etc. This issue can pose risks in sensitive clinical contexts, potentially undermining the credibility of ProFees. 

Therefore, in this paper, we have implemented a self-consistency strategy to enhance model output determinism. Specifically, our proposed model, the RCI-enhanced MDM classifier, is executed $K$ times concurrently, which avoids latency increases, with costs scaling proportionally. Each inference run generates intermediate predictions for all the MDM elements. Subsequently, a majority vote is conducted across the intermediate predictions of each element from these parallel runs. The consensus prediction for each element is then fed into the decision tree module to determine the MDM level and then the final CPT code. 

In cases where no majority is reached for an element, ties are broken by selecting the first prediction in the sorted result list from the parallel runs. For example, if the result list is \{1, 2, 2, 2, 3, 4, 4\}, the final output should be 2, while for \{1, 2, 2, 2, 4, 4, 4\}, 2 will still be chosen over 4 due to its precedence in tie-breaking.


\subsection{Decision Tree} 
This module considers outputs from both the Encounter Type Classifier and the RCI-enhanced MDM Classifier, along with additional EHR data, based on which specific MDM rules are applied to produce the final CPT E/M code.\footnote{An associated LLM justification of the MDM level prediction is also passed down from the MDM Classifier module and output together with the final E/M code.}

We do not release the full prompts \emph{verbatim} due to compliance and business confidentiality reasons. Instead, we share the prompt templates to capture the essence of our prompts to facilitate further research and community adoption. They can be found in Appendix~\ref{sec:prompts_appendix}.
\section{Data Gathering and Annotation}
\label{sec:data}


\begin{table}[t]
\caption{Datasets overview. \textit{Avg. \# Words} refers to the average number of words in input SOAP notes.}
\label{tab: Datasets}
\small
\centering
\begin{tabular}{@{}lrlr@{}}
\toprule
Dataset       & \# Samples    & Usage & Avg. \# Words \\ \midrule
Platinum          & 52        & dev   & 543                       \\
Disagreement      & 65        & dev   & 451                       \\
Test        & 99        & test  & 434                       \\ \bottomrule
\end{tabular}
\end{table}

\paragraph{Data Collection.}
We collected 216 real-world encounters from hospital production databases. The dataset comprises de-identified outpatient/preventative medicine encounters, each represented by a SOAP note, together with additional patient EHR data, such as demographics and medical orders.

Each encounter is annotated with a CPT E/M code assigned by the physician, along with a corresponding code provided by a professional coder. The coder's code either confirms or corrects the physician's selection and is used as the final billing code. However, the dataset lacks detailed justification for how the final CPT E/M code was determined, including the assessed complexity levels for the three MDM elements. Therefore, we engaged an internal medical expert to provide fine-grained annotations of both the E/M code and the MDM element complexity for each encounter.

These encounters were sampled incrementally to reflect the real-world distribution of CPT E/M codes observed in production data, using a larger dataset with coarse-level CPT labels as the reference. The sample spans multiple physician specialties, including family practice, internal medicine, urology, hematology, oncology, and cardiology (see Figures \ref{fig: dev_specialty_dist} and \ref{fig: test_specialty_dist}), rather than being limited to a single domain. Subsequently, our internal expert coders reviewed the set to confirm that the inclusiveness and coverage of specialties is sufficiently broad.
\paragraph{Data Annotation.}
For each encounter, our internal expert annotates the problem, data, and risk of complexity levels together with free-text justifications. The expert also assigned an overall CPT E/M code along with a justification for the selection. Using this procedure, we annotated a total of 117 encounters for model development. The CPT codes in the test set were annotated by the same expert in the same way as the development set.

Our analysis revealed a high degree of disagreement between the coders. Specifically, the expert agreed with the professional coder's CPT E/M codes on 52 encounters (44\%), and disagreed on the remaining 65 (56\%). Accordingly, we partitioned the development dataset into two subsets: \textbf{Platinum}, comprising encounters with agreement and thus more reliable ground-truth labels; and \textbf{Disagreement}, containing encounters where the coders diverged, reflecting more challenging coding decisions. Initial prompt tuning and critic-checklist design are performed exclusively on the \textit{Platinum} subset. We then carry out iterative error analysis on both \textit{Platinum} and \textit{Disagreement} with our internal expert, refining prompts design while correcting any labelling errors identified.

\paragraph{Final Datasets.}
Our final dataset comprises three subsets, as shown in Table \ref{tab: Datasets}. The \textbf{Platinum} and \textbf{Disagreement} subsets serve as our primary development data. The separate \textbf{Test} subset is used for model evaluation. Given the stringent compliance, data safety and privacy requirements associated with collecting real-world patient data, large-scale data acquisition is challenging. Thus, we aimed to include a diverse range of CPT E/M codes across these subsets. Distributions of the CPT E/M codes (See Figures~\ref{fig: platinum_code_dist},~\ref{fig: disagreement_code_dist} and \ref{fig: outpatient_code_dist}) and the data annotation interface (See Figure \ref{fig: annotation_interface}) can be found in Appendix~\ref{sec:datasets}.
\begin{table*}[htb]
\small
\centering
\caption{Evaluation results of accuracy improvements on the Test set. Results are averaged over 5 runs.}
\label{tab:results_covanent}
\begin{tabular}{lccccc}
\toprule
\textbf{Model} & \textbf{CPT Acc. (\%)} & \textbf{MDM Acc. (\%)} & \textbf{PC Acc. (\%)} & \textbf{DC Acc. (\%)} & \textbf{RC Acc. (\%)} \\
\midrule
System A & $\downarrow$ 3.12 ± 0.64 & $\downarrow$ 4.99 ± 0.40 & / & /& / \\
\midrule
Single prompt & --- ± 2.19 & --- ± 1.54 & --- ± 3.04 & --- ± 1.53 & --- ± 2.14 \\
Single prompt + CoT & $\uparrow$ 19.20 ± 1.79 & $\uparrow$ 13.26 ± 0.85 & $\uparrow$ 21.88 ± 1.93 & $\uparrow$ 15.06 ± 1.67 & $\uparrow$ 2.83 ± 1.44 \\
Single prompt + Full Info + CoT & $\uparrow$ 29.00 ± 2.19 & $\uparrow$ 22.36 ± 0.88 & $\uparrow$ 24.47 ± 0.83 & $\uparrow$ 22.82 ± 2.83 & $\uparrow$ 19.77 ± 0.64 \\
\midrule
ProFees (Zero-Shot) & $\uparrow$ 27.40 ± 1.00 & $\uparrow$ 24.45 ± 0.45 & $\uparrow$ 26.83 ± 0.00 & $\uparrow$ 27.21 ± 1.80 & $\uparrow$ 19.30 ± 1.18 \\
ProFees (Zero-Shot + RCI) & $\uparrow$ 29.73 ± 0.58 & $\uparrow$ 33.08 ± 0.60 & $\uparrow$ 32.32 ± 1.36 & $\uparrow$ 43.69 ± 1.36 & $\uparrow$ 23.22 ± 1.80 \\
ProFees (Few-Shot) & $\uparrow$ 29.73 ± 0.58 & $\uparrow$ 26.41 ± 0.82 & $\uparrow$ 28.39 ± 0.68 & $\uparrow$ 28.78 ± 1.80 & $\uparrow$ 22.04 ± 0.68 \\
ProFees (Few-Shot + RCI | Full) & $\uparrow$ 33.73 ± 0.58 & $\uparrow$ 33.99 ± 0.60 & $\uparrow$ 35.06 ± 0.00 & $\uparrow$ 39.76 ± 1.18 & $\uparrow$ 27.14 ± 0.68 \\
\bottomrule
\end{tabular}
\end{table*}

\section{Evaluation and Results}

\subsection{Experiment Settings}
\paragraph{Dataset.} We combine the \textbf{Platinum} and \textbf{Disagreement} subset, with a total of 117 encounters, and use them as the ``tuning'' dataset to refine prompts of both ProFees and the baselines. This ensures that both models are exposed to the same amount and type of generalizable knowledge, extracted by the internal expert from the error cases in the tuning set. The \textbf{Test} dataset, on the other hand, comprises the other 99 encounters. 

\paragraph{Vector Database.} We build the VDB with both the tuning/development and test datasets with the indexing procedures specified in Section~\ref{sec:dynamic_few_shot}. To ensure no information leakage occurs during the retrieval of dynamic few-shot exemplars from the VDB, we adopt the leave-one-out strategy, i.e., always filtering out the query's corresponding example from the retrieved list. Additionally, semantic dense retrieval is performed with $N=3$ for Top-$N$ exemplars retrieved (post leave-one-out filtering).

\paragraph{Baselines.}
We compare against (i) \textbf{System A}, a commercial rule-based coding tool,\footnote{The vendor name and detailed description are withheld to preserve anonymity.} and  
(ii) four \textbf{Single prompt} variants ranging from a single-prompt system to prompts augmented with CoT and full contextual information (i.e.\ EHR data).
Note that we use Azure \textbf{GPT-4o-2024-05-13} as the foundation LLM for both the baseline \textbf{Single Prompt} variants and our \textbf{Profees} model.

\paragraph{Evaluation Metrics.} We conduct accuracy-based evaluation at different granularity levels of classification. These include both the final CPT accuracy and that of the intermediate MDM element complexity levels. For each metric, we report the average result as well as the standard deviation over 5 runs with different random seeds. 
Note that the intermediate complexity prediction are not available for \textbf{System A} and are therefore omitted.

For each of the 5 runs, we set the number of predictions made for each test encounter to be $K=3$ to enforce the self-consistency majority vote (see Section~\ref{sec:self_consistency}), while balancing the cost and latency of LLM calling. In addition, we set the temperature for all LLM calls to be 0.

For setting of RCI round, we found empirically that ProFees's intermediate complexity prediction settled with 1 round of RCI for each MDM element. For simplicity, throughout the experiment, we also refer such 1-round setting as RCI.


\subsection{Overall Results}

Table~\ref{tab:results_covanent} presents performance, as measured by accuracy improvements over the ``Single prompt'' baseline, on the 99-encounter Test set.  
Our full model (\textit{Few-Shot + RCI}) achieves the highest CPT accuracy, 
36.85\% higher than the commercial System A, 
33.73\% higher than the Single prompt baseline, and 4.73\% over the strongest baseline (i.e. Single prompt + Full Info + CoT).  
Notably, adding RCI to our model lifts MDM accuracy by 4\% in the few-shot setting and 2.33\% in the zero-shot setting, confirming the value of structured self-critique.
Moreover, our full model significantly outperforms the strongest baseline on MDM, PC, DC and RC accuracy by 8--17 absolute points, further demonstrating its efficacy. Furthermore, a detailed ablation study can be found in Appendix~\ref{sec:ablation}. We also found that the overall trend of performance improvements brought by ProFees preserved on the internal-external coder agreement subset of the Test set, confirming that ProFees coding capability is generalizable towards unseen real-world encounters. Detailed results on the agreement subset can be found in Appendix~\ref{sec:additional_experiments}.


\section{Conclusion and Future Work}
This work presents the first systematic, production‐oriented study of automating CPT E/M coding with large language models. We propose a modular architecture that (i) retrieves context-relevant chain-of-thought exemplars, (ii) predicts both encounter type and MDM element complexities, (iii) performs RCI to critic those intermediate decisions, and (iv) consolidates results with deterministic rule logic and a Self-Consistency ensemble.   
On a de-identified, expert-annotated Test dataset of 99 encounters, our model Profees improves CPT accuracy by 4.73 \% over the strongest LLM baseline and by 30.61 \% over the commercial CPT E/M coding software 
System A.
Our future work includes extending the model to support multiple codes, CPT-modifiers and generating synthetic datasets for edge-case testing and enriching our VDB. 

\section*{Limitations}
ProFees currently predicts only one CPT E/M code per encounter, and we plan to support predicting multiple codes (e.g. for a preventive medicine visit, in which a problem-oriented service is also provided). It can be easily supported by extending the Encounter Type Classifier to output multiple encounter types. We also acknowledge the current lack of publicly available datasets for CPT E/M coding within the broader research community. To address this gap, we are curating and plan to release a synthetic dataset to facilitate research on improving model performance, robustness, and scalability in this domain.

\section*{Ethical Considerations}
While ProFees demonstrates a significant improvement in CPT E/M coding accuracy, its integration into EHR systems requires careful consideration of several key ethical issues to ensure its responsible deployment. 

\textbf{Privacy and data integrity.} Our development and evaluation efforts were conducted in a secure, HIPAA-compliant environment. As mentioned in Section~\ref{sec:data}, the development of ProFees utilize de-identified real-patient data. For deployment, ProFees will be deployed in a secure cloud environment with state-of-the-art safeguards to ensure patient confidentiality. 

\textbf{Accountability and liability.} Automating clinical coding may introduce ambiguity on accountability for coding errors. Ultimately, ProFees acts as an assistive tool, but not an autonomous decision-making tool, that helps physicians and professional coders, who take the final responsibility. 

\textbf{Algorithmic biases and fairness.} Our model was tuned and evaluated on a small-scale real-world dataset, which may contain latent biases. Thus, there is a risk that the model could learn (e.g.\ through few-shot examples) to systematically assign different CPT E/M codes based on patterns in encounter notes correlated with patient demographics (e.g.\ race and gender). To mitigate this risk, we are curating a synthetic dataset to enable targeted evaluation of model behaviors. 

\textbf{Interpretability and transparency.} Trust in a healthcare setting demands interpretability. While frontier LLMs are often considered blackboxes, we have developed ProFees to improve its interpretability. Specifically, through carefully designed prompts and architecture, we elicit sophisticated, multi-step reasoning from LLMs, leverage both internal and external clinical coding knowledge, and generate detailed rationales for coding outputs. This interpretability is not only crucial for gaining user trust and enabling human oversight, but also providing clear rationales for external audits. 

\textbf{Automation bias and human oversight.} ProFees' strong performance (a $>30$-point accuracy improvement) may create a risk of \emph{automation bias}. Specifically, it is plausible that human users might develop a degree of \emph{reliance} higher than what is warranted. ProFees generates detailed rationales that justify suggested codes. The UI/UX and workflow design should enable and facilitate human review of the codes and corresponding rationales. Moreover, continued user education of their roles and responsibility is necessary. 

\section*{Acknowledgement}
We would like to thank other members of Oracle Health \& AI for their collaboration while deploying ProFees in production, and Irfan Bulu, Raefer Gabriel, Neil Hauge, Mark Johnson, Kiran Rama, Amitabh Saikia, Vishal Vishnoi, and Krishnaram Kenthapadi for insightful feedback and discussions.

\bibliography{custom}

\appendix

\section{MDM Guidelines}\label{sec:mdm_guidelines}
Table~\ref{tab:mdm_levels} elaborates the MDM guidelines from the 2024 edition of the CPT Professional Manual book~\cite{ama2024cpt} that we have leveraged for building ProFees.

\begin{table*}[htbp]
\tiny
\centering
\caption{Levels of Medical Decision Making (MDM), extracted from the 2024 edition of MDM guidelines \cite{ama2024cpt}}
\label{tab:mdm_levels}

\begin{tabularx}{\textwidth}{
  >{\raggedright\arraybackslash}X X X X
}
\toprule
\textbf{Level of MDM} \begin{itemize}[nosep,leftmargin=*]
  \item Based on 2 out of 3 MDM Elements
\end{itemize} &
\textbf{Number and Complexity of Problems Addressed at the Encounter} &
\textbf{Amount and/or Complexity of Data Reviewed and Analyzed}\begin{itemize}[nosep,leftmargin=*]
  \item Each unique test, order, or document contribute to the combination of 2 or 3 in Cat. 1 below.
\end{itemize} &
\textbf{Risk of Complications and/or Morbidity or Mortality of Patient Management} \\
\midrule

\textbf{Straightforward} &\textbf{Minimal}
\begin{itemize}[nosep,leftmargin=*]
  \item \textbf{1} self‐limited or minor problem
\end{itemize} &
\textbf{Minimal or none} &
\textbf{Minimal risk of morbidity from additional diagnostic testing or treatment} \\
\midrule

\textbf{Low} &\textbf{Low}
\begin{itemize}[nosep,leftmargin=*]
  \item \textbf{2} or more self-limited or minor problems; \textbf{or}
  \item \textbf{1} stable chronic illness; \textbf{or}
  \item \textbf{1} acute, uncomplicated illness or injury; \textbf{or}
  \item \textbf{1} stable acute illness; \textbf{or}
  \item \textbf{1} acute, uncomplicated illness/injury requiring inpatient or observation level of care
\end{itemize} &
\textbf{Limited} –– at least \emph{1} out of 2 categories:%
\begin{itemize}[nosep,leftmargin=*]
  \item \textbf{Cat.\,1: Tests and documents} –– \textbf{Any combination of 2 from the following}:
  \begin{itemize}
  \item Review of prior external note(s) from each unique source;
  \item Review of the result(s) of each unique test;
  \item Ordering of each unique test
  \end{itemize} \textbf{or}
  \item \textbf{Cat.\,2: Assessment requiring an independent historian(s)} –– For the categories of independent interpretation of tests and discussion of management or test interpretation, see moderate or high.
\end{itemize} &
\textbf{Low risk from additional diagnostic testing or treatment} \\
\midrule
\textbf{Moderate} & \textbf{Moderate}
\begin{itemize}[nosep,leftmargin=*]
  \item \textbf{1} or more chronic illness with exacerbation,
        progression, or side-effects of treatment; \textbf{or} 
  \item \textbf{2} or more stable chronic illnesses; \textbf{or}
  \item \textbf{1} undiagnosed new problem with uncertain prognosis; \textbf{or}
  \item \textbf{1} acute illness with systemic symptoms; \textbf{or}
  \item \textbf{1} acute complicated injury
\end{itemize} &
\textbf{Moderate} –– at least \emph{1} out of 3 categories:%
\begin{itemize}[nosep,leftmargin=*]
  \item \textbf{Cat.\,1: Tests, documents, or independent historian(s) -- Any combination of 3 from the following:}
        \begin{itemize}
            \item Review of prior external
note(s) from each unique
source;
\item Review of the result(s) of
each unique test;
\item Ordering of each unique test;
\item Assessment requiring an independent historian(s)
\end{itemize} \textbf{or}
  \item \textbf{Cat.\,2: Independent interpretation of tests}
  \begin{itemize}
      \item Independent interpretation of a test performed by another physician or other qualified health care professional (not separately reported);
  \end{itemize} \textbf{or}
  \item \textbf{Cat.\,3: Discussion of management or test
        interpretation}
        \begin{itemize}
            \item Discussion of management or test interpretation with external physician or other qualified health care professional/appropriate source (not separately reported)
        \end{itemize}
\end{itemize} & \textbf{Moderate risk of morbidity from additional diagnostic testing or treatment} 
\begin{itemize}
    \item \textit{Examples only}:
    \begin{itemize}
        \item Prescription drug management
        \item Decision regarding minor surgery with identified patient or procedure risk factors
        \item Decision regarding elective major surgery without identified patient or procedure risk factors
        \item Diagnosis or treatment significantly limited by social determinants of health
    \end{itemize}
\end{itemize} \\
\midrule                         %
\textbf{High} & \textbf{High}
\begin{itemize}[nosep,leftmargin=*]
  \item \textbf{1} or more chronic illnesses with severe exacerbation, progression, or side effects of treatment; \textbf{or}
  \item \textbf{1} acute or chronic illness or injury that poses a threat to life or bodily function
\end{itemize} &
\textbf{Extensive} –– at least \emph{2} out of 3 categories:%
\begin{itemize}[nosep,leftmargin=*]
  \item \textbf{Cat.\,1: Tests, documents, or independent historian(s)} -- \textbf{Any combination of 3 from the following:}
  \begin{itemize}
      \item Review of prior external note(s) from each unique source;
      \item Review of the result(s) of each unique test;
      \item Ordering of each unique test;
      \item Assessment requiring an independent historian(s)
  \end{itemize} \textbf{or}
  \item \textbf{Cat.\,2: Independent interpretation of tests}
  \begin{itemize}
      \item Independent interpretation of a test performed by another physician or other qualified health care professional (not separately reported);
  \end{itemize} \textbf{or}
  \item \textbf{Cat.\,3: Discussion of management or test
        interpretation}
        \begin{itemize}
            \item Discussion of management or test interpretation with external physician or other qualified health care professional/appropriate source (not separately reported)
        \end{itemize}
\end{itemize} & \textbf{High risk of morbidity from additional diagnostic testing or treatment} 
\begin{itemize}
    \item \textit{Examples only}:
    \begin{itemize}
        \item Drug therapy requiring intensive monitoring for toxicity
        \item Decision regarding elective major surgery with identified patient or procedure risk factors
        \item Decision regarding emergency major surgery
        \item Decision regarding hospitalization or escalation of hospital-level care
        \item Decision not to resuscitate or to de-escalate care because of poor prognosis
        \item Decision regarding parenteral controlled substances
    \end{itemize}
\end{itemize} \\
\bottomrule
  \end{tabularx}
\end{table*}



\section{Ablation Studies}
\label{sec:ablation}

\paragraph{Dynamic Few-Shot CoT Prompting.}
Replacing the verbose prompt with three retrieved CoT examples increases CPT accuracy by 4\% relative to the zero-shot CoT variant, i.e. ProFees (Zero-Shot + RCI), and by 2.3\% when comparing ProFees (Few-Shot) to ProFees (Zero-Shot).

\paragraph{Recursive Criticism and Improvement (RCI).}
Adding the RCI stage boosts all three intermediate complexity scores, yielding a 7.58\% percentage-point gain in overall MDM accuracy and a 4\% gain in final CPT accuracy. This confirms that a dedicated post-hoc critic effectively detects and corrects the most common intermediate errors.

\begin{figure}[t]
  \centering
  \begin{subfigure}[t]{0.48\textwidth}
    \centering
    \includegraphics[width=\linewidth]{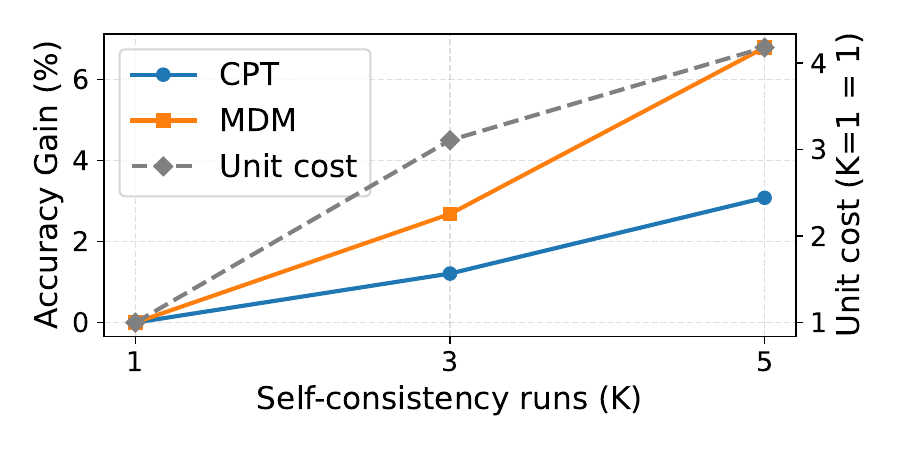}
    \caption{CPT and MDM accuracy gain and unit cost versus self-consistency runs.}
    \label{fig:cpt_mdm}
  \end{subfigure}
  \hfill
  \begin{subfigure}[t]{0.48\textwidth}
    \centering
    \includegraphics[width=\linewidth]{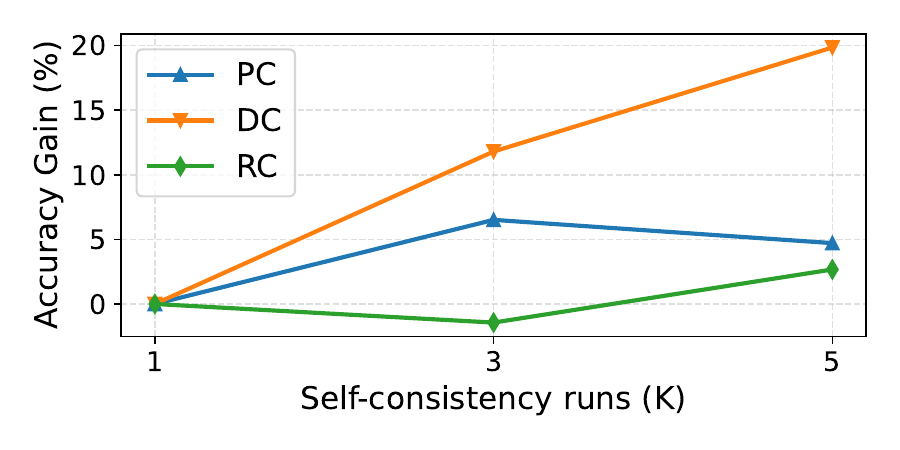}
    \caption{Intermediate accuracies gain (PC, DC, RC) versus self-consistency runs.}
    \label{fig:intermediate}
  \end{subfigure}
  \caption{Effect of self-consistency ($K$) on overall and intermediate performance.}
  \label{fig:self_consistency}
\end{figure}

\paragraph{Self-Consistency.} 
Figure~\ref{fig:self_consistency} reports \emph{relative} accuracy gains (compared to ProFees with 1 vote/self-consistency run, i.e. \(K{=}1\)), together with normalised cost. Increasing the vote count from one pass to \(K{=}3\) yields an additional \(+1.2\%\) CPT accuracy and \(+2.7\%\) MDM accuracy, but triples the unit cost.  
Raising the count further to \(K{=}5\) adds only \(+3.1\%\) CPT and \(+6.8\%\) MDM over the 1-vote ProFees while pushing the cost above a four-fold increase.  
Intermediate metrics show the following pattern: Data-Complexity gains \(+11.8\%\) at \(K{=}3\) and \(+19.9\%\) at \(K{=}5\); Problem-Complexity peaks at \(K{=}3\) then declines slightly, and Risk improves by just \(+2.7\%\) overall.  
To balance the accuracy improvements against the cost curve, we have chosen to fix \(K{=}3\) for all remaining experiments.

\section{High-Level Prompt Templates for CPT E/M Components}
\label{sec:prompts_appendix}

For compliance, privacy, and commercial reasons, we cannot release our full production prompts verbatim.  
Instead, we share \emph{template skeletons} that capture the \textbf{role}, \textbf{task}, and \textbf{output schema} of each LLM component while redacting institution-specific wording.

Each template specifies:
\begin{itemize}
    \item the \textbf{LLM role} (e.g., “Encounter-Type Classifier”, “Problem-Complexity Critic”);
    \item the \textbf{core task description} and minimal guidelines needed for replication;
    \item the \textbf{expected output format} (always JSON); and
    \item \textbf{placeholders} for dynamic inputs such as \verb|{{soap_note}}| or retrieved exemplars \verb|{{few_shot_examples}}|.
\end{itemize}

Researchers may adapt these templates to their own datasets and policies while remaining compatible with the architecture in the main paper.

\subsection{Encounter-Type Prompt}
{
\captionsetup[lstlisting]{labelformat=empty}
\begin{lstlisting}[language=, breakindent=0pt,basicstyle=\ttfamily\small, breaklines=true, showstringspaces=false, columns=fullflexible,frame=single]
prompt: |
    You are a medical coding assistant. Your task is to classify a patient's encounter into the appropriate CPT code category based on the provided SOAP note. The categories you need to consider are:

    <omitted encounter type with definition>

    Use the Medical Decision Making (MDM) guidelines to determine the correct category. Here is the information you have:

    - **SOAP Note**: {{soap_note}}
    {% if patient_type != '' %}
        - **Patient Type**: {{patient_type}} (New or Established)
    {% endif %}

    Based on this information, classify the encounter into one of the following categories:
    <omitted encounter type>
    
    Please provide the classification and a brief explanation for your decision.

    - **Output format**: {{format_instructions}}
    DO NOT ADD any content before or after the JSON.
\end{lstlisting}
}

\subsection{MDM-Complexity Prompt (Initial Pass)}

{
\captionsetup[lstlisting]{labelformat=empty}
\begin{lstlisting}[breakindent=0pt, basicstyle=\ttfamily\small, breaklines=true, columns=fullflexible, frame=single]
system: |
    You are a highly skilled medical coder specializing in CPT Evaluation and Management (E/M) codes. Your expertise lies in analyzing clinical documentation to assign accurate E/M codes while adhering strictly to AMA guidelines. Precision is critical, as any deviation in selecting the appropriate code based on the documented history, examination, and medical decision-making (MDM) can lead to claim rejections or audits. Your role ensures compliance, accuracy, and proper reimbursement for outpatient and consultation visits.

    ## Task:
    You will be given a patient encounter note together with relevant information about the visit (e.g. patient problems, medications and orders), and you are required to assess the medical decision making (MDM) elements of the encounter.

    ## Instructions:
    ### 1 - Assess Each Element of Medical Decision Making (MDM)

    There are three elements of MDM complexity, each with its own criteria:
    <omitted MDM guideline>
        
    ### 2 - Return each of the three MDM elements complexities together with your chain of thought reasoning for each based on the above criteria.

    {% if few_shot_examples%}
    ## Examples:
    Here are some examples:
    {{few_shot_examples}}
    
    {% endif %}
    ----------------------------
    ## IMPORTANT NOTES:
    <omitted MDM key notes>
    ----------------------------
    ## Output Format
    You should always return your output as a JSON object as per the schema specified by the user.
    Do not include any other text or formatting in your output. Only return the JSON object.
\end{lstlisting}
}
{
\captionsetup[lstlisting]{labelformat=empty}
\begin{lstlisting}[breakindent=0pt,basicstyle=\ttfamily\small,breaklines=true,columns=fullflexible,frame=single]
user: |
    ## Input:
    Here is the encounter note you need to analyze and classify:
    {{text}}
    {% if additional_info%}
    
    ## Additional Information:
    And here are the additional information about the visit:
    {{additional_info}}

    {% endif %}
    ## Output Format
    Please return the output as a JSON object following the below schema:
    {{format_instructions}}
    Do not include any other text or formatting in your output. Only return the JSON object.
\end{lstlisting}
}

\subsection{Critic Prompt for Problem Complexity}
{
\captionsetup[lstlisting]{labelformat=empty}
\begin{lstlisting}[breakindent=0pt, basicstyle=\ttfamily\small, breaklines=true, columns=fullflexible, frame=single]
prompt: |
    Now your role is a medical coding auditor with a keen eye for details.

    Your task is to review and reflect on the generated problem complexity assessment to ensure **accuracy, adherence to guidelines, and completeness**.

    # Instructions
    Review and reflect on the generated problem complexity assessment. Most of the time, it would be accurate, but you need to double-check that it adheres to the original guidelines as well as the additional instructions below.

    **Important Note:** Base all complexity assessments primarily on the 'Assessment and Plan' section, as it contains the provider's clinical judgment and final diagnosis.
    - Use the 'History of Present Illness' (HPI) only for contextual background, such as disease onset or duration for determining chronicity.
    - Do not rely on HPI to determine severity, uncertainty or progression, as it reflects the patient's subjective description, which may include exaggerations or inaccuracies.

    Apply System 2 Thinking:
    - Explicitly reason by breaking down your thought process step-by-step before reaching a conclusion.
    - The reasoning should mention why each instruction below (header, bullet and sub-bullet in this case) applies or does not apply to this encounter.
    - Ensure that you output the reasoning for **each** instruction.

    <omitted critic rules for problem complexity>

    # Output format:
    {{format_instructions}}
    ## Example output:
    The below is an example output to show the expected concise yet detailed reasoning and the expected output format. DO not mix up the conditions in the example output with the conditions in the note.

    DO NOT ADD any content before or after the JSON.
\end{lstlisting}
}

\subsection{Critic Prompt for Data Complexity}
{
\captionsetup[lstlisting]{labelformat=empty}
\begin{lstlisting}[breakindent=0pt, basicstyle=\ttfamily\small, breaklines=true, columns=fullflexible, frame=single]
prompt: |
    Now your role is a medical coding auditor with a keen eye for details. 

    Your task is to review and reflect on the generated data complexity assessment to ensure **accuracy, adherence to guidelines, and completeness**.
            
    # Instructions:
    Your role is to double-check the note and the data complexity in model response and ensure their compliance with the additional criteria below:

    Apply System 2 Thinking, meaning that for each instruction below, explicitly reason by breaking down your thought process step-by-step before reaching a conclusion.
    The reasoning should mention why each instruction (bullet in this case) was either followed, not followed, or not applicable. Ensure that you output the reason for each instruction.

    <omitted critic rules for data complexity>

    Finally return the final list of data complexity items with their types after your proposed changes, if any.

    # Output format: 
    {{format_instructions}}
    DO NOT ADD any content before or after the JSON.
\end{lstlisting}
}

\newpage
\subsection{Critic Prompt for Risk Assessment}
{
\captionsetup[lstlisting]{labelformat=empty}
\begin{lstlisting}[breakindent=0pt,basicstyle=\ttfamily\small, breaklines=true, columns=fullflexible, frame=single]
prompt: |
    ## **Key Risk Classification Criteria (Apply with Key Notes for General Risk Assessment)**
    1. xxx
    n. xxx

    # Every classification must be reviewed against **each** of the following **n key notes** to ensure accurate risk determination. These guidelines prevent errors caused by overlooked details, misinterpretation, or incorrect risk assignment.

    ## **Key Notes for General Risk Assessment (Must be Cross-Checked for Every Classification)**        
  
    <omitted critic rules for risk assessment>

    Apply **System 2 Thinking**, meaning that for each of the n instructions of Key Risk Classification Criteria, explicitly reason by breaking down your thought process step-by-step before reaching a conclusion. The reasoning should mention why each instruction was either **followed, not followed, or not applicable**, with exact references from the input SOAP Note and EHR information. Ensure that you output the reason and consider all instructions under **Key Notes for General Risk Assessment** for each evaluation.
    Your classification **must be cross-checked** against **each** of these n Key Notes to prevent misclassification due to oversight.
  
    <omitted key notes for risk assessment>

    ### **Classification Examples**
    The following examples illustrate the reasoning process required to do classifications. The chain of thought explicitly check each point of Official Guidelines and Criteria, while applies each key assessment rule to ensure an accurate classification.
  
    # **Output Format:**
    {{format_instructions}}

    DO NOT ADD any content before or after the JSON.
\end{lstlisting}
}

\newpage
\section{Datasets}
\label{sec:datasets}

This section presents more details of our datasets, including the CPT E/M code distribution over the three subsets and an example data annotation page.

\subsection{Distribution of CPT E/M Codes}

\begin{figure}[th]
\centering
\includegraphics[width=\linewidth]{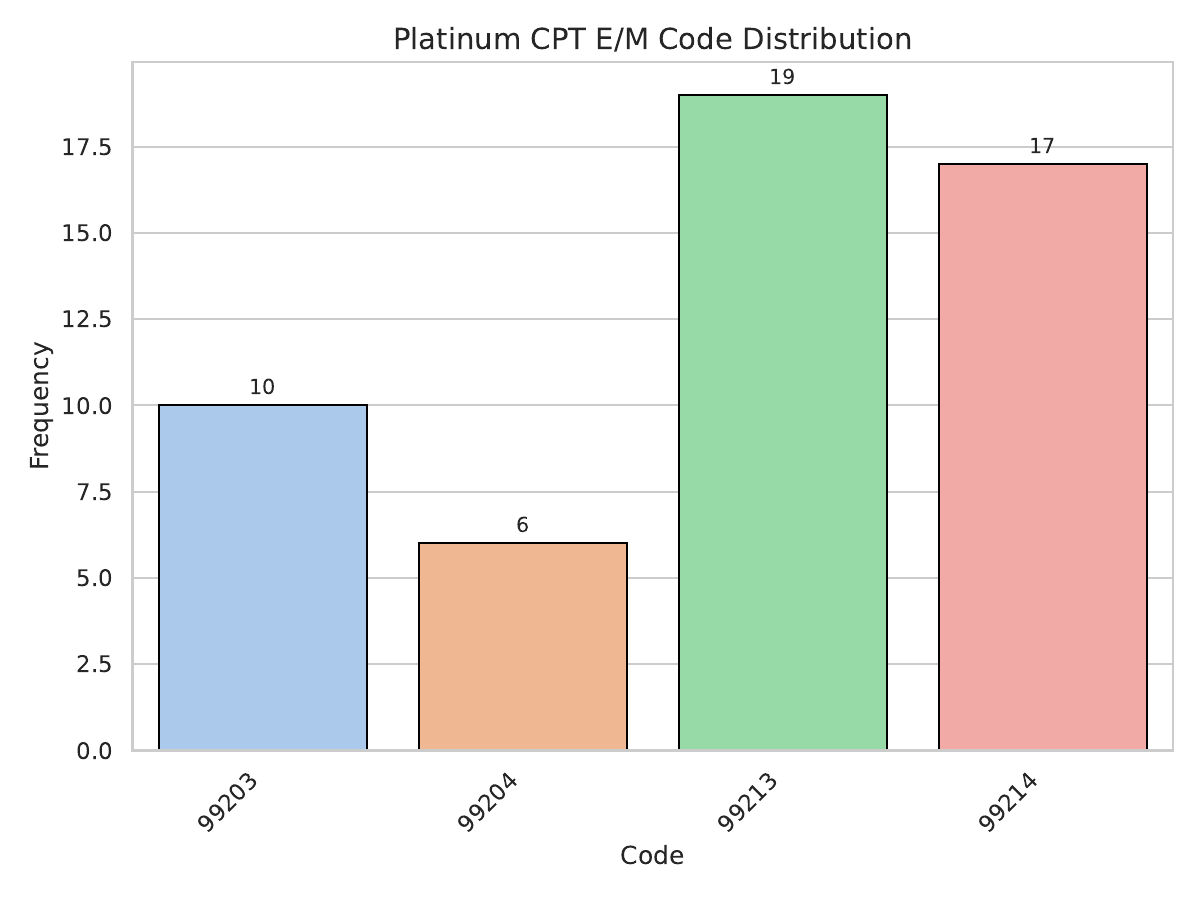}
\caption{Frequency distribution of CPT E/M codes on the Platinum subset.}
\label{fig: platinum_code_dist}
\end{figure}

\begin{figure}[th]
\centering
\includegraphics[width=\linewidth]{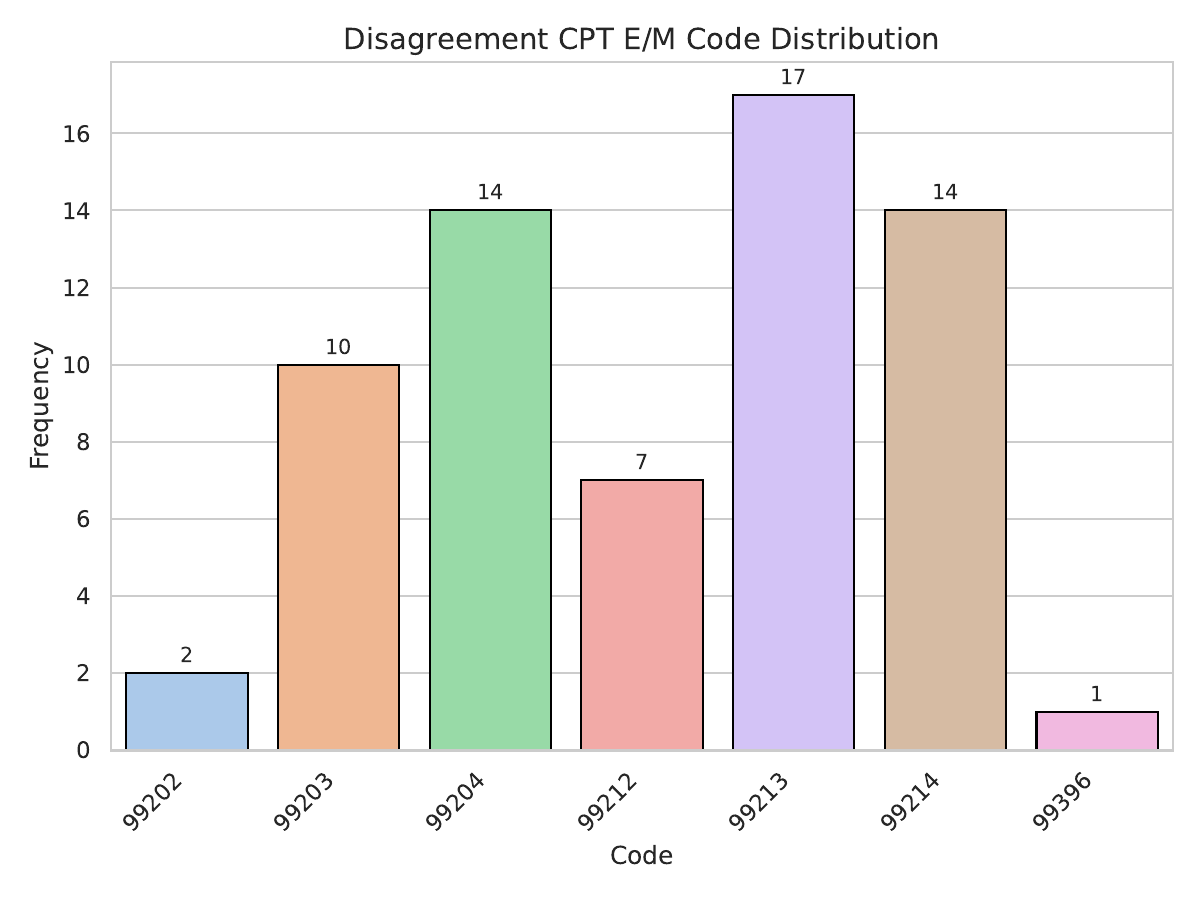}
\caption{Frequency distribution of CPT E/M codes on the Disagreement subset.}
\label{fig: disagreement_code_dist}
\end{figure}

\begin{figure}[th]
\centering
\includegraphics[width=\linewidth]{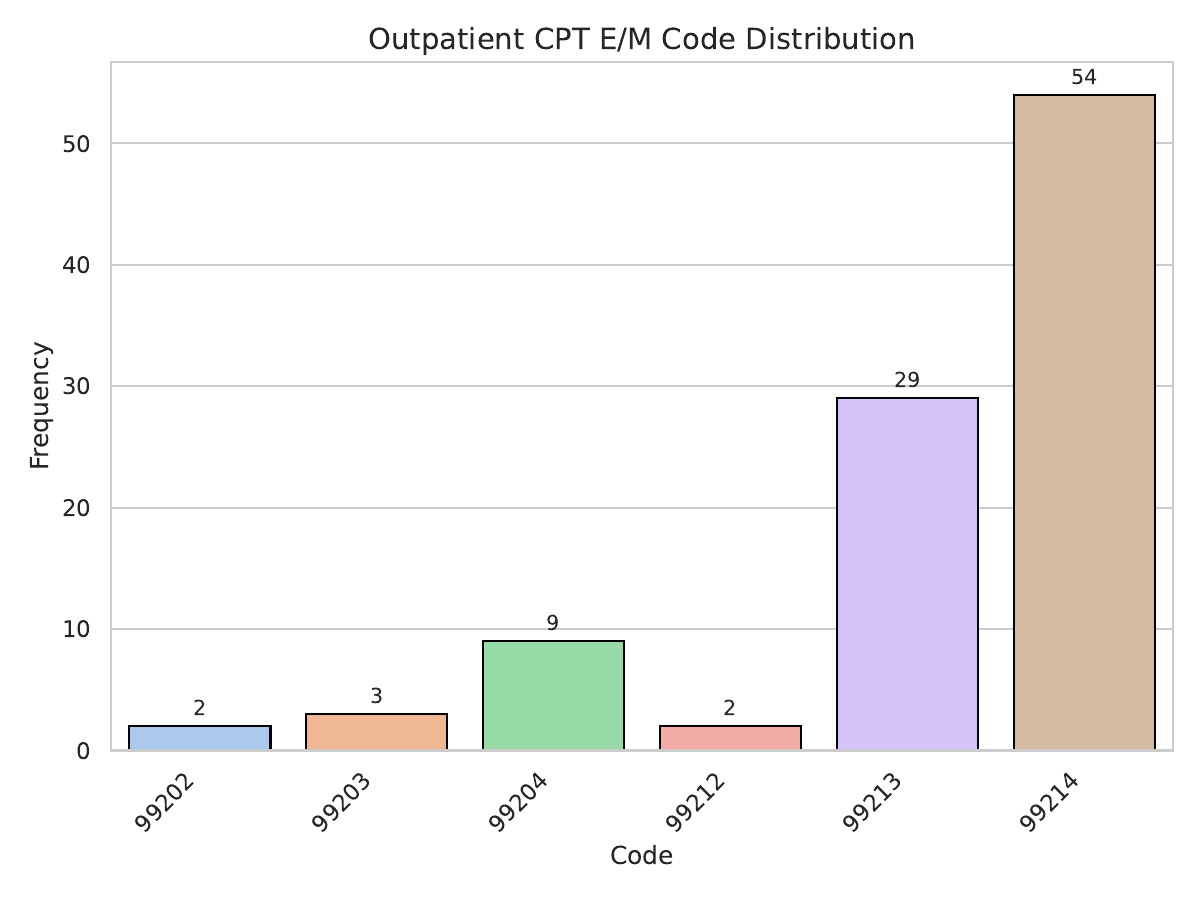}
\caption{Frequency distribution of CPT E/M codes on the Test subset.}
\label{fig: outpatient_code_dist}
\end{figure}

Figures~\ref{fig: platinum_code_dist}, \ref{fig: disagreement_code_dist} and \ref{fig: outpatient_code_dist} illustrate the distribution of CPT E/M codes in the \emph{Platinum}, \emph{Disagreement} and \emph{Test} datasets respectively.

\subsection{Specialty Distribution}
\begin{figure}[th]
\centering
\includegraphics[width=\linewidth]{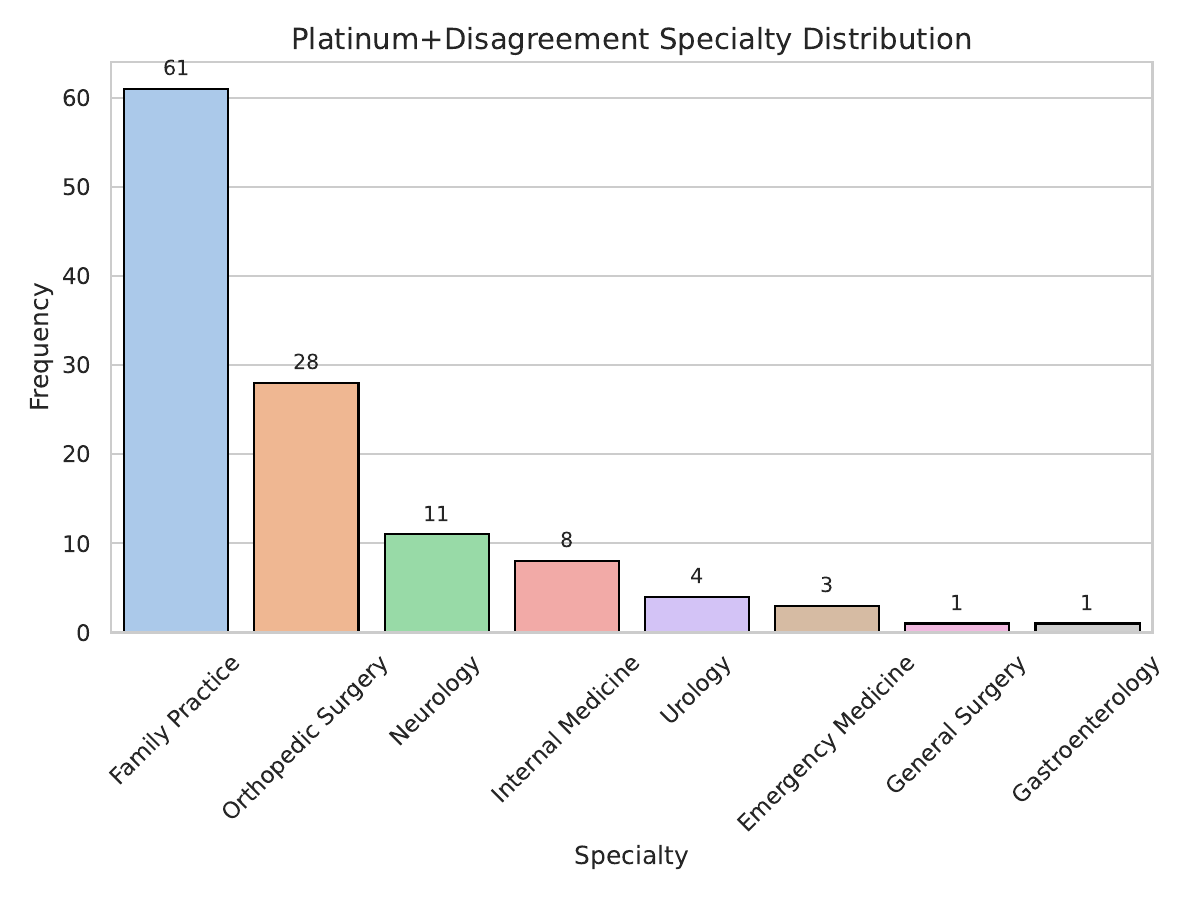}
\caption{Frequency distribution of physician specialties on the combined Platinum and Disagreement subset.}
\label{fig: dev_specialty_dist}
\end{figure}

\begin{figure}[th]
\centering
\includegraphics[width=\linewidth]{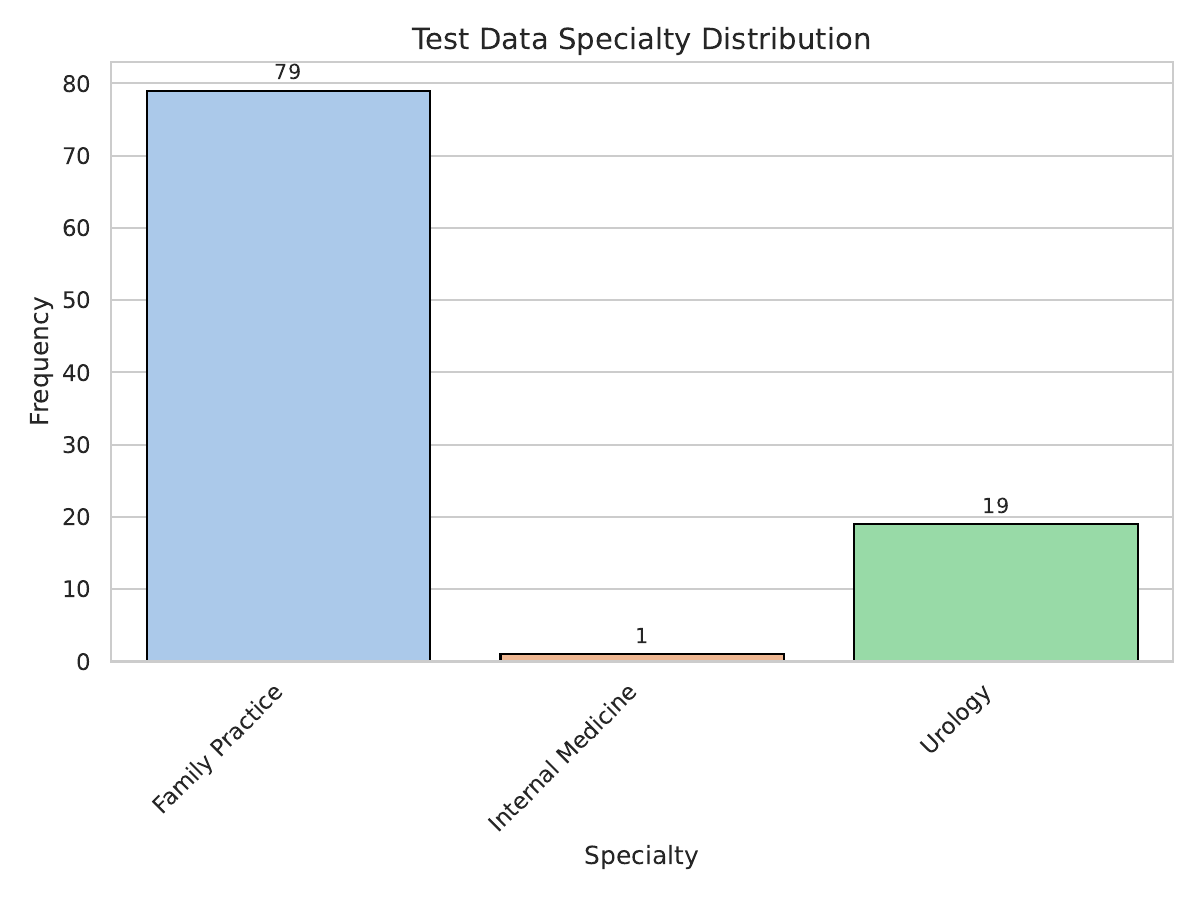}
\caption{Frequency distribution of physician specialties on the Test subset.}
\label{fig: test_specialty_dist}
\end{figure}

Figures~\ref{fig: dev_specialty_dist} and ~\ref{fig: test_specialty_dist} illutrate the distribution of physician specialties in the combined \emph{Platinum} + \emph{Disagreement}, and \emph{Test} datasets respectively.

\subsection{Data Annotation Interface}
\begin{figure*}[th]
\centering
\includegraphics[width=0.92\linewidth]{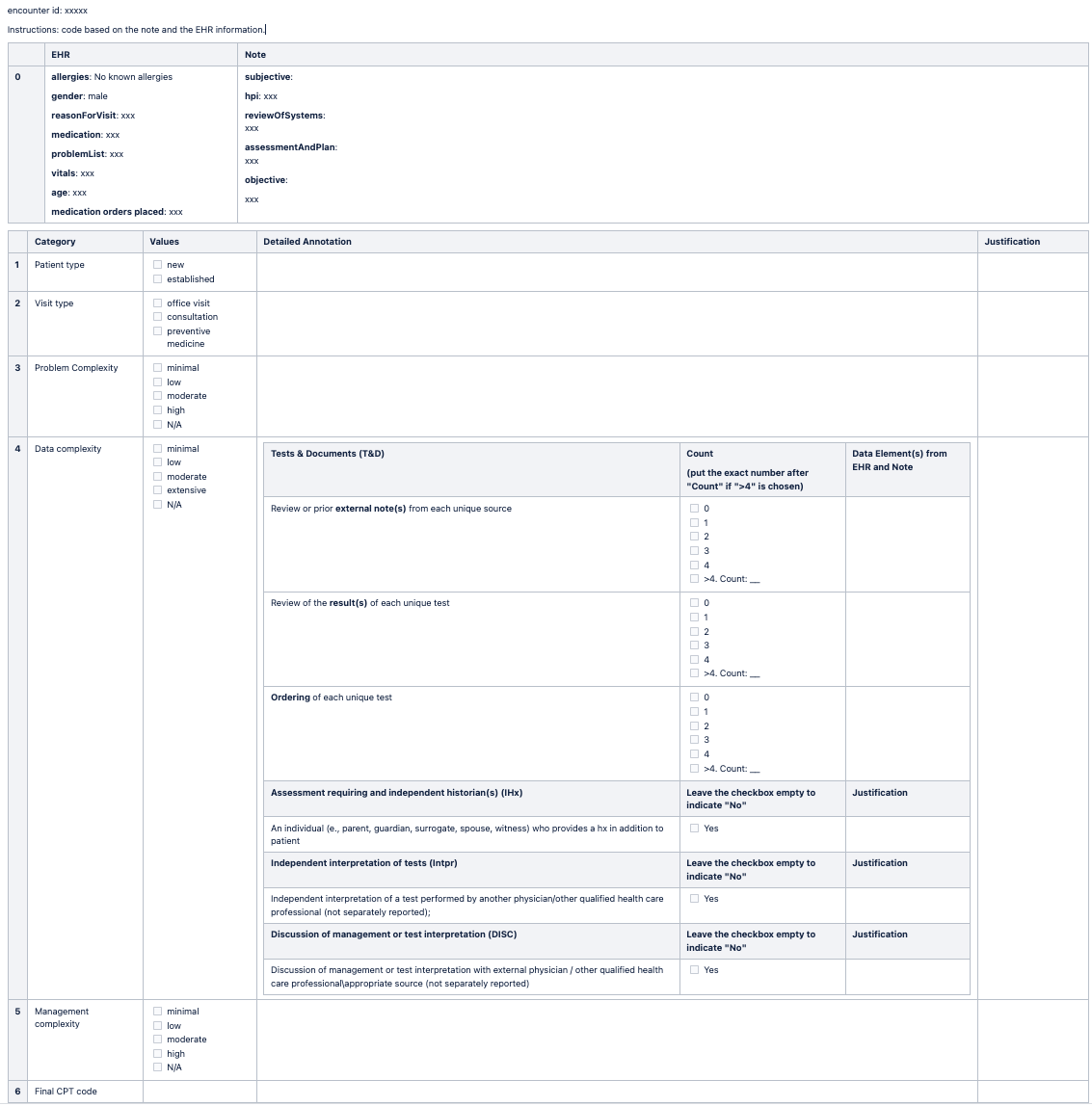}
\caption{Our data annotation interface.}
\label{fig: annotation_interface}
\end{figure*}

Figure \ref{fig: annotation_interface} shows our data annotation interface.

\clearpage
\section{Additional Experiments}\label{sec:additional_experiments}
We present the performance of ProFees on the agreement subset of our Test dataset in Table~\ref{tab:results_covanent_agreement_subset}, on which our internal coder agrees with external coders' CPT E/M codes. Specifically, they agree on 54 out of the 99 test encounters.

\begin{table*}[t]
\small
\centering
\caption{Evaluation of accuracy improvements on the agreement subset of Test set. Results are averaged over 5 runs.}
\label{tab:results_covanent_agreement_subset}
\begin{tabular}{lccccc}
\toprule
\textbf{Model} & \textbf{CPT Acc. (\%)} & \textbf{MDM Acc. (\%)} & \textbf{PC Acc. (\%)} & \textbf{DC Acc. (\%)} & \textbf{RC Acc. (\%)} \\
\midrule
ProFees (Zero-Shot) & $\uparrow$ 25.01 ± 1.07 & $\uparrow$ 26.95 ± 1.07 & $\uparrow$ 23.69 ± 0.00 & $\uparrow$ 26.94 ± 1.07 & $\uparrow$ 20.96 ± 1.07 \\
ProFees (Zero-Shot + RCI) & $\uparrow$ 30.65 ± 1.07 & $\uparrow$ 32.51 ± 1.07 & $\uparrow$ 32.94 ± 0.00 & $\uparrow$ 44.84 ± 3.21 & $\uparrow$ 24.04 ± 1.85 \\
ProFees (Few-Shot) & $\uparrow$ 26.33 ± 1.85 & $\uparrow$ 28.19 ± 1.85 & $\uparrow$ 25.54 ± 1.85 & $\uparrow$ 31.26 ± 1.07 & $\uparrow$ 25.28 ± 2.83 \\
ProFees (Few-Shot + RCI | Full) & $\uparrow$ 33.74 ± 0.00 & $\uparrow$ 35.59 ± 0.00 & $\uparrow$ 32.33 ± 1.07 & $\uparrow$ 41.14 ± 1.85 & $\uparrow$ 27.75 ± 0.00 \\
\bottomrule
\end{tabular}
\end{table*}

\section{Synthetic Example}
\label{sec: synthetic example}
This section shows an example of synthetically generated data, aligning with the data used in our work.


\begin{table}[H]
    {\small
    \begin{tabular}{@{}ll@{}}
        \textbf{Patient Type:} & New \\
        \textbf{Encounter Type:} & Office or Outpatient Service \\
        \textbf{CPT Code:} & 99203 \\
        \textbf{Problem Complexity:} & Low \\
        \textbf{Data Complexity:} & Straightforward \\
        \textbf{Risk of Complications:} & Low \\
        \textbf{MDM Complexity:} & Low \\
    \end{tabular}
    }
\end{table}

\newpage
{
\captionsetup[lstlisting]{labelformat=empty}
\begin{lstlisting}[breakindent=0pt, basicstyle=\ttfamily\small, breaklines=true, columns=fullflexible, frame=single,caption={Synthetic SOAP note.}]
SUBJECTIVE

Chief Complaint: Left ear pain and mild swelling after swimming.

History of Present Illness:
Emily Carter is a 24-year-old new patient presenting today with discomfort in her left ear that started yesterday after a swim at her local pool. She describes the pain as a dull ache, rated 4/10, accompanied by mild swelling around the ear canal. She states, "My ear feels sore and a little puffy since I went swimming, but I don't feel sick otherwise." She denies fever, chills, or drainage from the ear. She also mentions some redness and itchiness in both eyes recently after spending time outdoors, but clarifies this was much worse a few days ago and is now stable since starting antihistamine eye drops. No vision changes or other new symptoms. "My eyes were super itchy and red a few days back, but the drops really helped." No other complaints.

REVIEW OF SYSTEMS
ENT: Positive for left ear pain and mild swelling; negative for hearing loss, discharge, or tinnitus.
Eyes: Mild residual itching; no redness or vision changes today.
Constitutional: Denies fever, chills, malaise.

OBJECTIVE
Physical Exam:
- General: Alert, comfortable, no acute distress. Vitals stable.
- HEENT: Left external ear canal erythematous, mildly edematous, tender. No purulent discharge. TM intact, no erythema. Right ear normal. Conjunctivae clear, oropharynx clear.
- Neck: No lymphadenopathy.

ASSESSMENT AND PLAN
- Acute otitis externa, left ear, likely from swimming.
    - Recommend acetic acid 2% otic solution (OTC) for canal acidification.
    - Keep ear dry; avoid water until resolved.
    - Pain control with acetaminophen or ibuprofen as needed.
    - Return for worsening (fever, redness, persistent pain).
    - No antibiotics needed at this time.
    - If no improvement in 48-72 hrs or symptoms worsen, re-evaluate for prescription therapy/workup.

\end{lstlisting}
}


\end{document}